\documentclass[journal]{IEEEtran}
\usepackage{graphicx}
\usepackage{enumerate}
\usepackage{amsmath,amsfonts,amssymb}
\usepackage{color}
\usepackage{multirow}
\usepackage[noend]{algpseudocode}
\usepackage{algorithmicx,algorithm}
\usepackage{setspace}
\usepackage{lineno,hyperref}
\usepackage{booktabs}
\usepackage[table]{xcolor}
\modulolinenumbers[5]
\usepackage{etoolbox}

\begin{document}

\title{SuperYOLO: Super Resolution Assisted Object Detection in Multimodal Remote Sensing Imagery}

\author{Jiaqing Zhang,
	Jie Lei,~\IEEEmembership{Member,~IEEE},
	Weiying Xie,~\IEEEmembership{Member,~IEEE},
	Zhenman Fang,~\IEEEmembership{Member,~IEEE},
	Yunsong Li,~\IEEEmembership{Member,~IEEE},
	and Qian Du,~\IEEEmembership{Fellow,~IEEE}
\thanks{This work was supported in part by the National Natural Science Foundation of China under Grant 62071360. (Corresponding~authors: Jie Lei)

Jiaqing Zhang, Jie Lei, Weiying Xie, Yunsong Li are with the State Key
Laboratory of Integrated Services Networks, Xidian University, Xi’an 710071,
China (e-mail: jqzhang\underline{ }2@stu.xidian.edu.cn; jielei@mail.xidian.edu.cn; wyxie@xidian.edu.cn; ysli@mail.xidian.edu.cn).

Zhenman Fang is with School of Engineering Science, Simon Fraser University, Burnaby, BC, Canada (e-mail: zhenman@sfu.ca).

Qian Du is with the Department of Electronic and Computer Engineering, Mississippi State University, Starkville, MS 39759 USA (e-mail: du@ece.msstate.edu).}

}

\markboth{IEEE TRANSACTIONS ON GEOSCIENCE AND REMOTE SENSING,~Vol.~X, No.~X, 2023}%
{Zhang \MakeLowercase{\textit{et al.}}: SuperYOLO: Super Resolution Assisted Object Detection in Multimodal Remote Sensing Imagery}

\maketitle

\begin{abstract}  
\textcolor{blue}{This is the pre-acceptance version, to read the final version please go to IEEE TRANSACTION ON GEOSCIENCE AND REMOTE SENSING on IEEE Xplore.} Accurately and timely detecting multiscale small objects that contain tens of pixels from remote sensing images (RSI) remains challenging. 
Most of the existing solutions primarily design complex deep neural networks to learn strong feature representations for objects separated from the background, which often results in a heavy computation burden. In this paper, we propose an accurate yet fast object detection method for RSI, named SuperYOLO, which fuses multimodal data and performs high resolution (HR) object detection on multiscale objects by utilizing the assisted super resolution (SR) learning and considering both the detection accuracy and computation cost. First, we utilize a symmetric compact multimodal fusion (MF) to extract supplementary information from various data for improving small object detection in RSI. Furthermore, we design a simple and flexible SR branch to learn HR feature representations that can discriminate small objects from vast backgrounds with low-resolution (LR) input, thus further improving the detection accuracy. Moreover, to avoid introducing additional computation, the SR branch is discarded in the inference stage and the computation of the network model is reduced due to the LR input. Experimental results show that, on the widely used VEDAI RS dataset, SuperYOLO achieves an accuracy of 75.09\% (in terms of $\text{mA}{{\text{P}}_{\text{50}}}$), which is more than 10\% higher than the SOTA large models such as YOLOv5l, YOLOv5x and RS designed YOLOrs. Meanwhile, the parameter size and GFOLPs of SuperYOLO are about 18x and 3.8x less than YOLOv5x. Our proposed model shows a favorable accuracy-speed trade-off compared to the state-of-art models. The code will be open sourced at \url{https://github.com/icey-zhang/SuperYOLO}.
\end{abstract}
\section{Introduction}
\IEEEPARstart{O}{bject} detection plays an important role in various fields involving computer-aided diagnosis or autonomous piloting. Over the past decades, numerous excellent deep neural network (DNN) based object detection frameworks \cite{6909475, 7410526, redmon2016you, 8099809, ren2016faster} have been proposed, updated, and optimized in computer vision. The remarkable accuracy enhancement of DNN-based object detection frameworks owes to the application of large-scale natural datasets with accurate annotations \cite{5206848, lin2014microsoft, everingham2010pascal}. 

\begin{figure*}[htbp]
	\centering
	\includegraphics[scale=0.9]{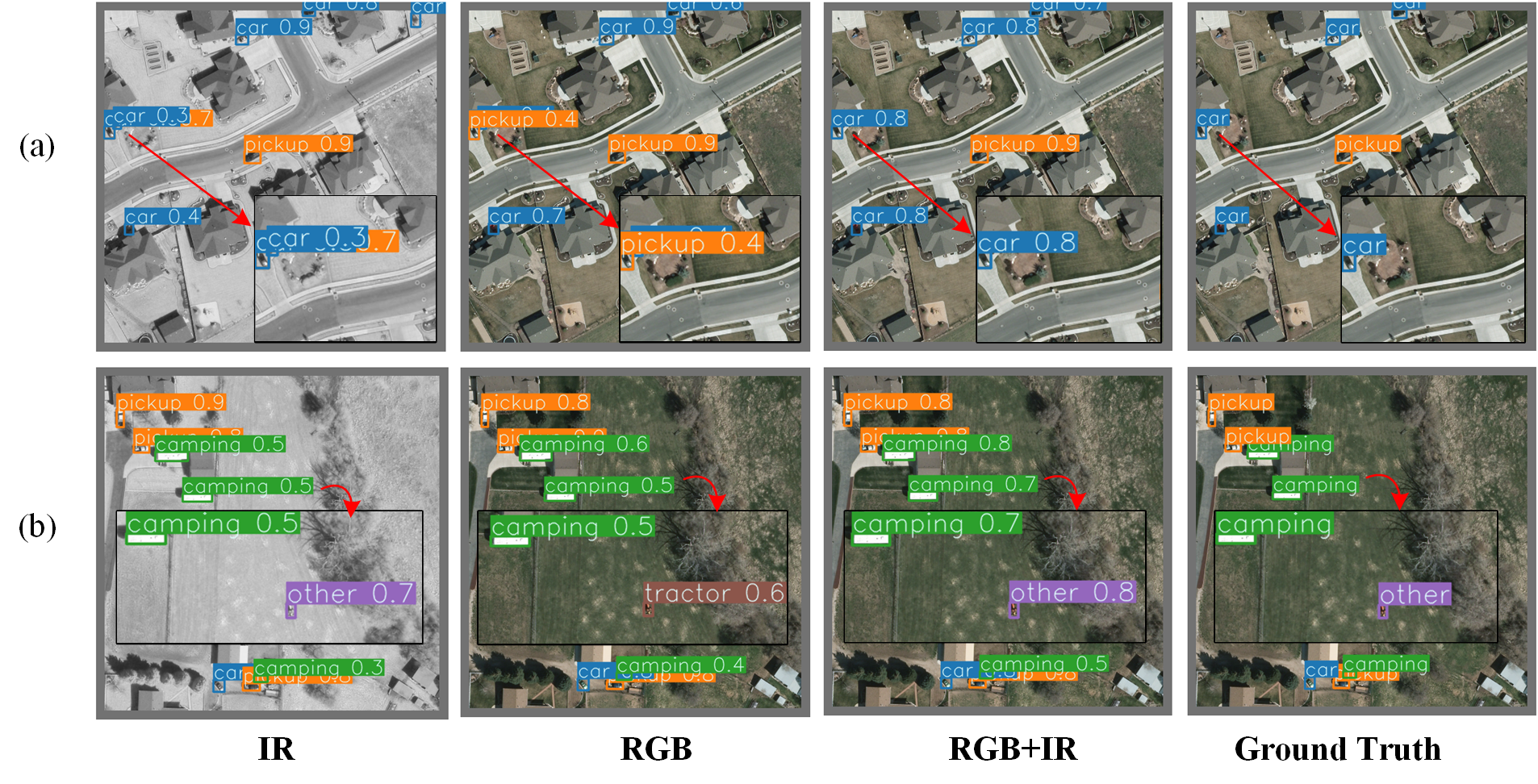}
	\centering
    \vspace{-0.1in}
	\caption{Visual comparison of RGB image, IR image, and ground truth (GT). The IR image provides vital complementary information for resolving the challenges in RGB detection. The object car in (a) is considerably small within a vast area. In (b), the objects have large-scale variation, to which the scale of a car is smaller than that of a camping vehicle. The fusion of RGB and IR modalities effectively enhances detection performance.}
	\vspace{-0.1in}
	\label{difficult}
\end{figure*}

Compared with natural scenarios, there are several vital challenges for accurate object detection in remote sensing images (RSIs).
First, the number of labeled samples is relatively small, which limits the training of DNNs to achieve high detection accuracy.
Second, the size of objects in RSI is much smaller, accounting for merely tens of pixels in relation to the complicated and broad backgrounds \cite{zheng2020foreground, pang2019mathcal}. Moreover, the scale of those objects is diverse with multiple categories \cite{deng2018multi}. As shown in Fig. \ref{difficult} (a), the object car is considerably small within a vast area.  As shown in Fig. \ref{difficult} (b), the objects have large-scale variations, to which the scale of a car is smaller than that of a camping vehicle. 

Currently, most object detection techniques are solely designed and applied for a single modality such as RGB and Infrared (IR) \cite{8953881, 7480356}. Consequently, with respect to object detection, its capability to recognize objects on the earth's surface remains insufficient due to the deficiency of complementary information between different modalities \cite{9174822}. As imaging technology flourishes, RSIs collected from multimodality become available and provide an opportunity to improve detection accuracy. For example, as shown in Fig. \ref{difficult}, the fusion of two different multimodalities (RGB and IR) can effectively enhance the detection accuracy in RSI. Sometimes the resolution of one modality is low which requires technique to improve the resolution to enhance information. Recently, super resolution technology has shown great potential in remote sensing fields  \cite{wang2020ultra,razzak2023multi,jiang2019edge,xiao2021satellite}. Benefiting from the vigorous development of the convolutional neuron network (CNN), the resolution of the remote sensing image has achieved high texture information to be interpreted. However, due to the high computation cost of the CNN network, the application of the SR network in real-time practical tasks has become a hot topic in current research.


In this study, our motivation is to \textit{propose an on-board real-time object detection framework for multimodal RSIs to achieve high detection accuracy and high inference speed without introducing additional computation overhead.} Inspired by recent advances in real-time compact neural network models, we choose small-size YOLOv5s \cite{yolov5} structure as our detection baseline. It can reduce deployment costs and facilitate rapid deployment of the model. Considering the high resolution (HR) retention requirements for small objects, we remove the Focus module in the baseline YOLOv5s model, which not only benefits defining the location of small dense objects but also enhances the detection performance. Considering the complementary characteristics in different modalities, we propose a multimodal fusion (MF) scheme to improve the detection performance for RSI. We evaluate different fusion alternatives (pixel-level or feature-level) and choose pixel-level fusion for low computation cost. 

Lastly and most importantly, we develop a super resolution (SR) assurance module to guide the network to generate HR features that are capable of identifying small objects in vast backgrounds, thereby reducing false alarms induced by background-contaminated objects in RSI. Nevertheless, a naive SR solution can significantly increase the computation cost. Therefore, we set the auxiliary SR branch engaged in the training process and remove it in the inference stage, facilitating spatial information extraction in HR without increasing computation cost.

In summary, this paper makes the following contributions.

\begin{itemize}
	
    
    \item We propose a computation-friendly pixel-level fusion method to combine inner information bi-directionally in a symmetric and compact manner. It efficiently decreases the computation cost without sacrificing accuracy compared with feature-level fusion.
   
    \item We introduce an assisted SR branch into multimodal object detection for the first time. Our approach not only makes a breakthrough in limited detection performance but also paves a more flexible way to study outstanding HR feature representations that are capable of discriminating small objects from vast backgrounds with LR input.
    
   \item  Considering the demand for high-quality results and low-computation cost, the SR module functioning as an auxiliary task is removed during the inference stage without introducing additional computation. 
    The SR branch is general and extensible and can be inserted in the existing fully convolutional network (FCN) framework.
	
	\item The proposed SuperYOLO markedly improves the performance of object detection, outperforming SOTA detectors in real-time multimodal object detection. 
	Our proposed model shows a favorable accuracy-speed trade-off compared to the state-of-art models.
	
\end{itemize}


\begin{figure*}
	\centering
	\includegraphics[scale=0.8]{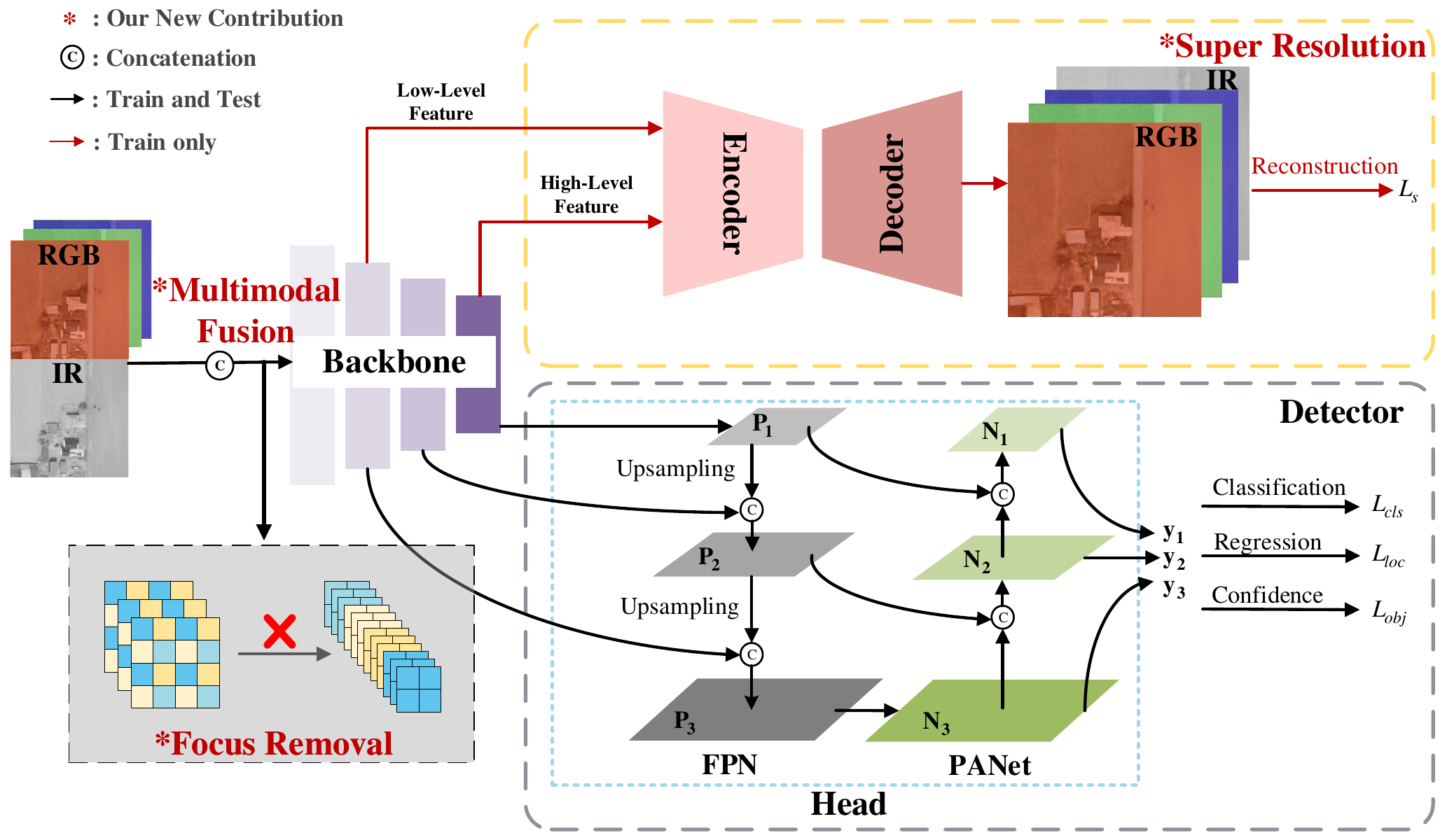}
	\centering
	\caption{The overview of the proposed SuperYOLO framework. Our new contributions include 1) removal of the Focus module to reserve high resolution, 2) multimodal fusion, and 3) assisted SR branch. The architecture is optimized in terms of Mean Square Error (MSE) loss for the SR branch and task-specific loss for object detection. During the training stage, the SR branch guides the related learning of the spatial dimension to enhance the high resolution information preservation for the backbone. During the test stage, the SR branch is removed to accelerate the inference speed equal to the baseline.}
	\vspace{-0.1in}
	\label{framework}
\end{figure*}
\section{Related Work}
\label{sec:Related Work}

\subsection{Object Detection with Multimodal Data}
Recently, multimodal data has been widely leveraged in numerous practical application scenarios, including visual question answering \cite{zhang2021multimodal}, auto-pilot vehicles \cite{chen2021multimodal}, saliency detection \cite{CHEN2021107740}, and remote sensing classification \cite{zhu2021spatial}. It is found that combining the internal information of multimodal data can efficiently transfer complementary features to avoid certain information of a single modality from being omitted.

In the field of RSI processing, there exist various modalities (e.g., Red-Green-Blue (RGB), Synthetic Aperture Radar (SAR), Light Detection and Ranging (LiDAR), Infrared (IR), panchromatic (PAN) and multispectral (MS) images) from diverse sensors, which can be fused with complementary characteristics to enhance the performance of various tasks \cite{sun2021deep, li2022asymmetric, gao2021hyperspectral}.
For example, the additional IR modality \cite{9273212} captures longer thermal wavelengths to improve the detection under difficult weather conditions. Manish \textit{et al.} \cite{9273212} proposed a real-time framework for object detection in multimodal remote sensing imaging, in which the extended version conducted mid-level fusion and merged data from multiple modalities. Despite that multi-sensor fusion can enhance the detection performance as shown in Fig \ref{difficult}, hardly can its low-accuracy detection performance and to-be-improved computing speed meet the requirements of real-time detection tasks.

The fusion methods are primarily grouped into three strategies, i.e., pixel-level fusion, feature-level fusion, and decision-level fusion methods \cite{7182258}. The decision-level fusion methods fuse the detection results during the last stage, which may consume enormous computation resources due to repeated calculations for different multimodal branches. In the field of remote sensing, feature-level fusion methods are mainly adopted with multi branches. The multimodal images will be input into the parallel branches to extract respective independent features of different modalities, and then these features will be combined by some operations, such as attention module or simple concatenation. The parallel branches bring repeated computation as the modalities increase, which is not friendly in the real-time tasks in remote sensing.

In contrast, the adoption of pixel-level fusion methods can reduce unnecessary computation. In this paper, our proposed SuperYOLO fuses the modalities at the pixel-level to significantly reduce the computation cost and design operations in spatial and channel domains to extract inner information in the different modalities which can help enhance the detection accuracy.

\subsection{Super Resolution in Object Detection}
In recent literature, the performance of small object detection can be improved by multi-scale feature learning \cite{lin2017feature, li2020density}, context-based detection \cite{chen2017r}. These methods always enhance the information representation ability of the network in different scales but ignore the high-resolution contextual information reservation. Conducted in a pre-processing step, SR has proven to be effective and efficient in various object detection tasks \cite{noh2019better, haris2018task}. Shermeyer \textit{et al.} \cite{shermeyer2019effects} quantified its effect on the detection performance of satellite imaging by multiple resolutions of RSI. Based on generative adversarial networks (GANs), Courtrai \textit{et al.} \cite{courtrai2020small} utilized SR to generate HR images, which were fed into the detector to improve its detection performance.
Rabbi \textit{et al.} \cite{rabbi2020small} leveraged a Laplacian operator to extract edges from the input image to enhance the capability of reconstructing HR images, thus improving its performance in object localization and classification.
Hong \textit{et al.} \cite{ji2019vehicle} introduced a cycle-consistent GAN structure as an SR network and modified faster R-CNN architecture to detect vehicles from enhanced images that are produced by the SR network.
In these works, the adoption of the SR structure has effectively addressed the challenges regarding small objects. However, compared with single detection models, additional computation is introduced, which attributes to the enlarged scale of the input image by HR design.

Recently, Wang \textit{et al.} \cite{9157434} proposed an SR module that can maintain HR representations with LR input while reducing the model computation in segmentation tasks. Inspired by the \cite{9157434}, we design an SR assisted branch. In contrast to the aforementioned work in which the SR is realized in the start stage, the assisted SR module guides the learning of high-quality HR representations for the detector, which not only strengthens the response of small dense objects but also improves the performance of the object detection in spatial space. Moreover, the SR module is removed in the inference stage to avoid extra computation.

\begin{figure*}
	\centering
	\includegraphics[scale=1]{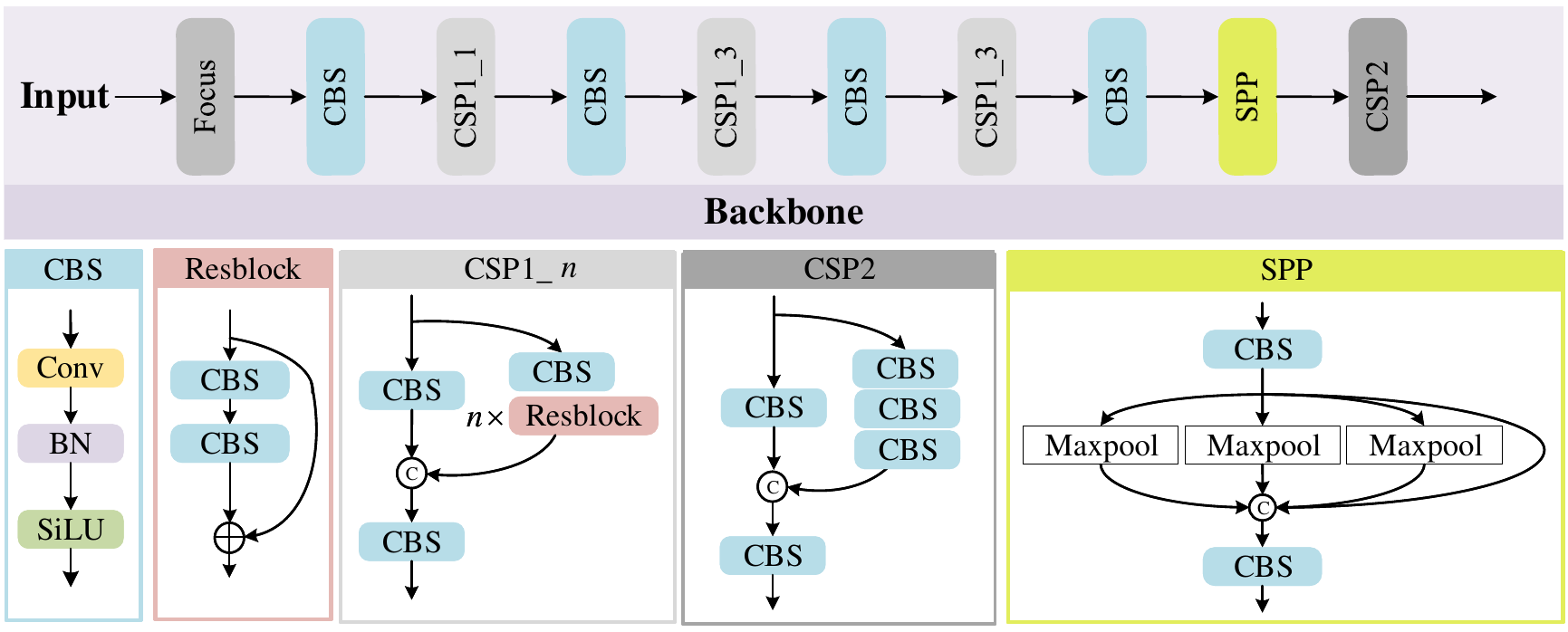}
	\caption{The backbone structure of YOLOv5s. The low-level texture and high-level semantic features are extracted by stacked CSP, CBS, and SPP structures.}
	\vspace{-0.1in}
	\label{backbone}
\end{figure*}


\section{Baseline Architecture}
\label{sec:Baseline YOLOv5s Architecture}

As shown in Fig. \ref{framework}, the baseline YOLOv5 network consists of two main components: the Backbone and Head (including the Neck). The backbone is designed to extract low-level texture and high-level semantic features. Next, these hint features are fed to Head to construct the enhanced feature pyramid network from top to bottom to transfer robust semantic features and from bottom to top to propagate a strong response of local texture and pattern features. This resolves the various scale issue of the objects by yielding an enhancement of detection with diverse scales.



\begin{figure*}[htpb]
	\centering
	\includegraphics[scale=1.1]{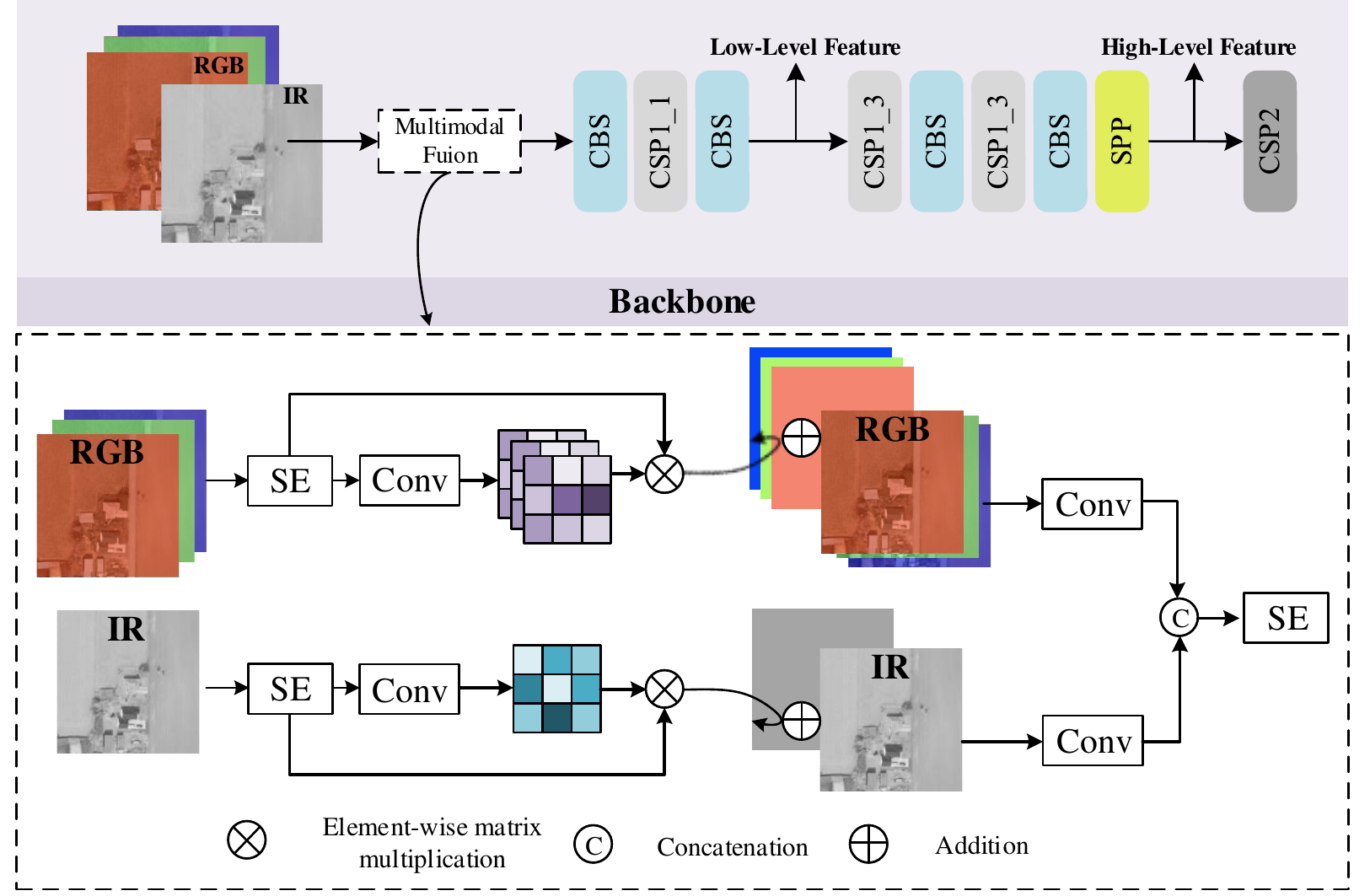}
	\centering
	\caption{The architecture of the multimodal fusion (MF) module at the pixel level.}
	\vspace{-0.1in}
	\label{Mfusion}
\end{figure*}


In Fig. \ref{backbone}, CSPNet \cite{9150780} is utilized as the Backbone to extract the feature information, consisting of numerous sample Convolution-Batch-normalization-SiLu (CBS) components and Cross Stage Partial (CSP) modules. The CBS is composed of operations of convolution, batch normalization, and activation function SiLu \cite{elfwing2018sigmoid}. The CSP duplicates the feature map of the previous layer into two branches and then halves the channel numbers through $1\times 1$ convolution, by which the computation is therefore reduced. With respect to the two copies of the feature map, one is connected to the end of the stage, and the other is sent into ResNet blocks or CBS blocks as the input. Finally, the two copies of the feature map are concatenated to combine the features, which is followed by a CBS block. 
The SPP (Spatial Pyramid Pooling) module \cite{he2015spatial} is composed of parallel Maxpool layers with different kernel sizes and is utilized to extract multiscale deep features. The low-level texture and high-level semantic features are extracted by stacked CSP, CBS, and SPP structures.

\noindent \textbf{Limitation 1:} It is worth mentioning that the Focus module is introduced to decrease the number of computations. As shown in Fig. \ref{framework} (bottom left), inputs are partitioned into individual pixels and reconstructed at intervals and finally concatenated in the channel dimension. The inputs are resized to a smaller scale to reduce the computation cost and accelerate the network training and inference speed. However, this may sacrifice object detection accuracy to a certain extent, especially for small objects vulnerable to resolution.

\noindent \textbf{Limitation 2:} It is known that the backbone of YOLO employs deep convolutional neural networks to extract hierarchical features with a stride step of 2, through which the size of the extracted features is halved. Hence, the feature size retained for multiscale detection is far smaller than that of the original input image. For example, when the input image size is 608, the sizes of output features for the last detection layer are 76, 38, and 19, respectively. LR features may result in the missing of some small objects.

\section{SuperYOLO Architecture}
\label{sec:Our SuperYOLO Architecture}

As summarized in Fig. \ref{framework}, we introduce three new contributions to our SuperYOLO network architecture. First, we remove the Focus module in the Backbone and replace it with an MF module, to avoid resolution degradation and thus accuracy degradation. Second, we explore different fusion methods and choose the computation-efficient pixel-level fusion to fuse RGB and IR modalities to refine dissimilar and complementary information. Finally, we add an assisted SR module in the training stage, which reconstructs the HR images to guide the related Backbone learning in spatial dimension and thus maintain HR information. In the inference stage, the SR branch is discarded to avoid introducing additional computation overhead.

\subsection{Focus Removal}
\label{Sec:Focus Removal}

As presented in Section~\ref{sec:Baseline YOLOv5s Architecture} and Fig. \ref{framework} (bottom left), the Focus module in the YOLOv5 backbone partitions images at intervals on the spatial domain and then reorganize the new image to resize the input images. Specifically, this operation is to collect a value for every group of pixels in an image and then reconstruct it to obtain smaller complementary images. The size of the rebuilt images decreases with the increase in the number of channels. 
As a result, it causes resolution degradation and spatial information loss for small targets. Considering that the detection of small targets depends more heavily on higher resolution, the Focus module is abandoned and replaced by an MF module (shown in Fig. \ref{Mfusion}) to prevent the resolution from being degraded.

\subsection{Multimodal Fusion} 
\label{subsec:fusion}

The more information is utilized to distinguish objects, the better performance can be achieved in object detection. Multimodal fusion is an effective path for merging different information from various sensors. The decision-level, feature-level, and pixel-level fusions are the three mainstream fusion methods that can be deployed at different depths of the network. Since decision-level fusion requires enormous computation, it is not considered in SuperYOLO. 

\begin{figure*}[htpb]
	\centering
	\includegraphics[scale=1]{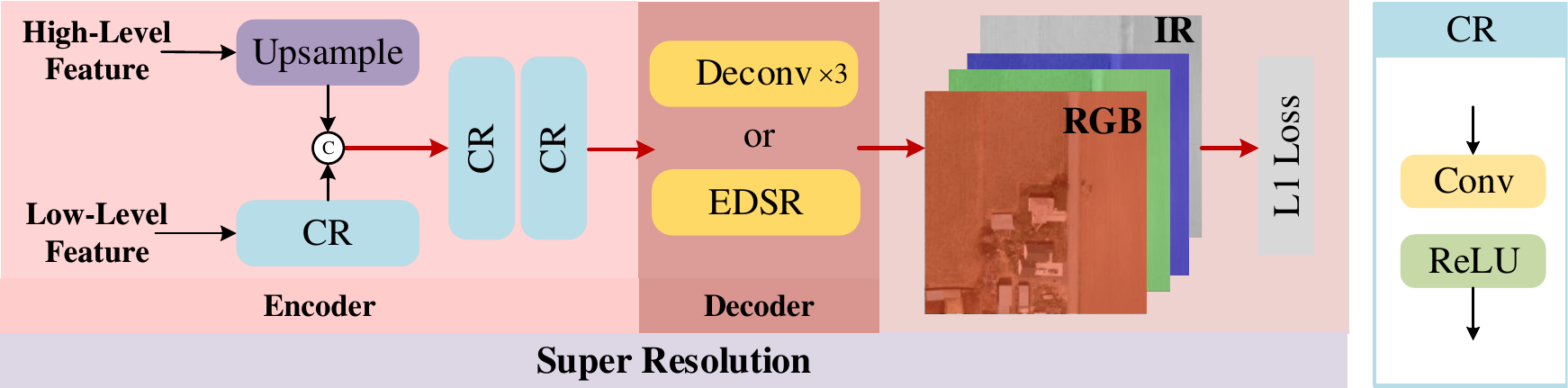}
	\centering
	\caption{The super resolution (SR) structure of SuperYOLO. The SR structure can be regarded as a simple Encode-Decoder model. The low-level and high-level features of the backbone are selected to fuse local textures patterns and semantic information, respectively.}
	\vspace{-0.1in}
	\label{SR}
\end{figure*}

We propose a pixel-level multimodal fusion (MF) to extract the shared and special information from the different modalities. The MF can combine multimodal inner information bi-directionally in a symmetric and compact manner. As shown in Fig. \ref{Mfusion}, for the pixel-level fusion, we first normalize an input RGB image and an input IR image into two intervals of $[0,1]$. And the input modalities $X_{RGB} , X_{IR} \in {\mathbb{R}}^{C \times H \times W}$ are subsampled  to $I_{RGB},I_{IR} \in {{\mathbb{R}}^{C\times \frac{H}{n}\times \frac{W}{n}}}$ which are fed to SE  blocks extracting inner information in channel domain \cite{hu2018squeeze} to generate $F_{RGB},F_{IR}$:
\begin{equation}
	F_{RGB}=SE(I_{RGB}),F_{IR}=SE(I_{IR}),
\end{equation}
Then the attention map that reveals the inner relationship of the different modalities in the spatial domain is defined as:
\begin{equation}
		m_{IR} = f_1(F_{IR}),
		m_{RGB} = f_2(F_{RGB}),
\end{equation}
where $f_1$ and $f_2$ represent $1 \times 1$ convolutions for the RGB and IR modalities, respectively. 
Here, $\otimes$ denotes element-wise matrix multiplication. 
Inner spatial information between the different modalities is produced by:
\begin{equation}
	F_{in1} = m_{RGB}  \otimes F_{RGB},F_{in2} = m_{IR}  \otimes F_{IR}.
\end{equation}
To incorporate internal inner-view information and spatial texture information, the features are added by the original input modalities and then fed into $1 \times 1$ convolutions. That the full features are:
\begin{equation}
	F_{ful1} = f_3(F_{in1} +I_{RGB}),F_{ful2} = f_4(F_{in2} + I_{IR}).
\end{equation}
where $f_3$ and $f_4$ represent $1 \times 1$ convolutions. Finally, the features are fused by:
\begin{equation}
	F_{o} = SE(Concat(F_{ful1},F_{ful2})).
\end{equation}
where $Concat(\cdot )$ denotes the concatenation operation along the channel axis. 
The result is then fed to the Backbone to produce multi-level features.
Note that, $X$ is subsampled to $1/n$ size of the original image to accomplish the SR module discussed in Section \ref{sec: Super Resolution} and to accelerate the training process. The $X$ represents the RGB or IR modality, and the sampled image is denoted as ${I} \in {{\mathbb{R}}^{C\times \frac{H}{n}\times \frac{W}{n}}}$ and generated by:
\begin{equation}
	I=D(X),
\end{equation}
where $D (\cdot)$ represents $n$ times downsampling operation using bilinear interpolation.

\begin{figure*}[htpb]
	\centering
	\includegraphics[scale=0.69]{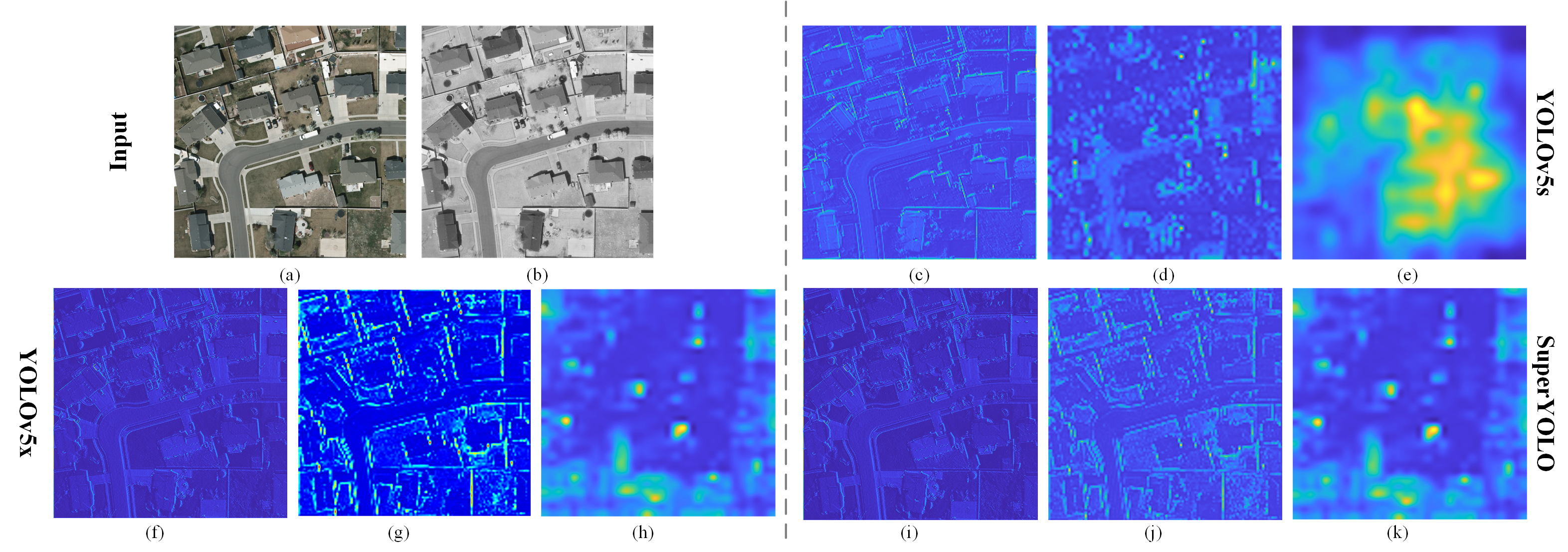}
	\centering
	\vspace{-0.05in}
	\caption{Feature-level visualization of backbone for YOLOv5s, YOLOv5x and SuperYOLO with the same input: (a) RGB input, (b) IR input; (c), (d), and (e) are the features of YOLOv5s; (f), (g), and (h) are the features of YOLOv5x; (i), (j) and (k) are the features of SuperYOLO. The features are upsampled to the same scale as the input image for comparison. (c), (f) and (i) are the features in the first layer. (d), (g) and (j) are the low-level features. (e), (h) and (k) are the high-level features in layers at the same depth.}
	\vspace{-0.1in}
	\label{feature}
\end{figure*}

\subsection{Super Resolution}
\label{sec: Super Resolution}
As mentioned in Section \ref{sec:Baseline YOLOv5s Architecture}, the feature size retained for multiscale detection in the backbone is far smaller than that of the original input image. Most of the existing methods conduct upsampling operations to recover the feature size. Unfortunately, this approach has produced limited success due to the information loss in texture and pattern, which explains that it is inappropriate to employ this operation to detect small targets that require HR preservation in RSI.

To address this issue, as shown in Fig. \ref{framework}, we introduce an auxiliary SR branch.
First, the introduced branch shall facilitate the extraction of HR information in the backbone and achieve satisfactory performance. Second, the branch should not add more computation to reduce the inference speed. It shall realize a trade-off between accuracy and computation time during the inference stage. Inspired by the study of Wang \textit{et al.} \cite{9157434} where the proposed super resolution succeeded in facilitating segmentation tasks without additional requirements, we introduce a simple and effective branch named SR into the framework. Our proposal can improve detection accuracy without computation and memory overload, especially under circumstances of LR input.

Specifically, the SR structure can be regarded as a simple Encode-Decoder model. We select the backbone's low-level and high-level features to fuse local textures and patterns and semantic information, respectively. As depicted in Fig. \ref{Mfusion}, we select the result of the fourth and ninth modules as the low-level and high-level features, respectively. The Encoder integrates the low-level feature and high-level feature generated in the backbone.  As illustrated in Fig. \ref{SR}, in Encoder, the first CR module is conducted on the low-level feature. For the high-level feature, we use an Upsampling operation to match the spatial size of the low-level feature and then we use a concatenation operation and two CR modules to merge the low-level and high-level features. The CR module includes a convolution and ReLU.
For the Decoder, the LR feature is upscaled to the HR space in which the SR module's output size is twice larger than that of the input image. As illustrated in Fig. \ref{SR}, the Decoder is implemented using three deconvolutional layers. 
The SR guides the related learning of spatial dimension and transfers it to the main branch, thereby improving the performance of object detection. In addition, we introduce EDSR \cite{lim2017enhanced} as our Encoder structure to explore the SR performance and its influence on detection performance. 

To present a more visually interpretable description, we visualize the features of backbones for YOLOv5s, YOLOv5x and SuperYOLO in Fig. \ref{feature}. The features are upsampled to the same scale as the input image for comparison. By comparing the pairwise images of (c), (f) and (i); (d), (g) and (j); (e) (h) and (k) in Fig. \ref{feature}, it can be observed that SuperYOLO contains clearer object structures with higher resolution with the assistance of the SR. Eventually, we obtain a bumper harvest in high-quality HR representation with the SR branch and utilize the Head of YOLOv5 to detect small objects.

\subsection{Loss Function}
The overall loss of our network consists of two components: detection loss $L_o$ and SR construction loss $L_s$, which can be expressed as
\begin{equation}
{{L}_{total}}={{c}_{1}}{{L}_{o}}+{{c}_{2}}{{L}_{s}},
\end{equation}
where ${c}_{1}$ and ${c}_{2}$ are the coefficients for a balance of the two training tasks.
The L1 loss (rather than L2 loss) \cite{zhao2016loss} is used to calculate the SR construction loss ${L}_{s}$ between the input image $X$ and SR result $S$, to which the expression is written as
\begin{equation}
{{L}_{s}}=\left\| S-X \right\|_{1}.
\end{equation}
The detection loss involves three components \cite{yolov5}: loss of judging whether there is an object ${L}_{obj}$, loss of object location ${L}_{loc}$, and loss of object classification ${L}_{cls}$, which are used to evaluate the loss of the prediction as
\begin{equation}
\label{L_o}
L_o={{\lambda }_{loc}}\sum\limits_{l=0}^{2}{{{a}_{l}}{{L}_{loc}}}+{{\lambda }_{obj}}\sum\limits_{l=0}^{2}{{{b}_{l}}}{{L}_{obj}}+{{\lambda }_{cls}}\sum\limits_{l=0}^{2}{{{c}_{l}}}{{L}_{cls}}.
\end{equation}

Here, equation \ref{L_o}, $l$ represents the layer of the output in head, $a_l$, $b_l$, and $c_l$ are the weights of different layers for the three loss functions, the weights ${{\lambda}_{loc}}$, ${{\lambda }_{obj}}$, and ${{\lambda }_{cls}}$ regulate error emphasis among box coordinates, box dimensions, objectness, no-objectness and classification.

\section{Experimental Results}
\label{sec:Experiment}

\subsection{Dataset}
The popular Vehicle Detection in Aerial Imagery (VEDAI) dataset  \cite{razakarivony2016vehicle} is used in the experiments, which contains cropped images obtained from the much larger Utah Automated Geographic Reference Center (AGRC) dataset. Each image collected from the same altitude in AGRC has approximately $16,000\times 16,000$ pixels, with a resolution of about $ 12.5 cm \times 12.5 cm $ per pixel. RGB and IR are the two modalities for each image in the same scenes. The VEDAI dataset consists of 1246 smaller images that focus on diverse backgrounds involving grass, highway, mountains, and urban areas. All images are in the size of $1024\times 1024$ or $512 \times 512 $. The task is to detect 11 classes of different vehicles such as car, pickup, camping, and truck. 



\subsection{Implementation Details}
Our proposed framework is implemented in PyTorch and runs on a workstation with an NVIDIA 3090 GPU. The VEDAI dataset is used to train our SuperYOLO. Following \cite{9273212}, the VEDAI dataset is devised to 10 fold cross-validation. In each split 1089 images are used for training and another 121 images are used for testing. The ablation experiments are conducted on the first fold of data, while the comparisons with previous methods are performed on the 10 folds by averaging their results. The annotations for each object in the image contain the coordinates of the bounding box center, the orientation of the object concerning the positive $x$-axis, the four corners of the bounding box, the class ID, a binary flag identifying whether an object is occluded, and another binary flag identify whether an object is cropped. We do not consider classes with fewer than 50 instances in the dataset, such as plane, motorcycle, and bus. So the annotations of the VEDAI dataset are converted to YOLOv5 format, and we transfer the ID of the interested class to $0, 1, ..., 7$, i.e., $N=8$. Then the center coordinates of the bounding box are normalized and the absolute coordinate is transformed into a relative coordinate. Similarly, the length and width of the bounding box are normalized to $[0,1]$. To realize the SR assisted branch, the input images of the network are downsampled from $1024 \times 1024$ size to $512 \times 512$ during the training process. In the test process, the image size is $512 \times 512$, which is consistent with the input of other algorithms compared. In addition, data is augmented with Hue Saturation Value (HSV), multi-scale, translation, left-right flip, and mosaic. The augmentation strategy is canceled in the test stage. The standard Stochastic Gradient Descent (SGD) \cite{bottou2010large} is used to train the network with a momentum of 0.937, weight decay of 0.0005 for the Nesterov accelerated gradients utilized, and a batch size of 2. The learning rate is set to 0.01 initially. The entire training process involves 300 epochs.

\subsection{Accuracy Metrics}
The accuracy assessment measures the agreements and differences between the detection result and the reference mask. The recall, precision, and mAP (mean Average Precision) are used as accuracy metrics to evaluate the performance of the methods to be compared with. The calculations of the precision and recall metrics are defined as
\begin{equation}
	Precision=\frac{TP}{TP+FP}
\end{equation}
\begin{equation}
	Recall=\frac{TP}{TP+FN}.
\end{equation}
\begin{table}[htpb]
	\small
	\centering
	\setlength{\tabcolsep}{1.5mm}{
		\caption{The Comparison Results of Model Size and Inference Ability in Different Baseline YOLO Frameworks on the First Fold of the VEDAI Validation Set.}
		\label{modelsize}
		\begin{tabular}{c|c|c|c|c}
			\toprule[1.2pt]
			\textbf{Method}  & \textbf{Layers} $\downarrow$ & \textbf{Params} $\downarrow$& \textbf{GFLOPs} $\downarrow$&  \textbf{$\text{mA}{{\text{P}}_{\text{50}}}$} $\uparrow$ \\
			\midrule
			YOLOv3  & 270   & 61.5M  & 52.8  & 62.6    \\
			YOLOrs  & 241   & 20.2M  & 46.4  & 55.8  \\
			YOLOv4  & 393  & 52.5M   & 38.2  & \textbf{65.7}   \\
			YOLOv5s & \textbf{224}    & \textbf{7.1M}          & \textbf{5.32}  & 62.2   \\
			YOLOv5m & 308  & 21.1M  & 16.1  & 64.5  \\
			YOLOv5l & 397& 46.6M & 36.7  & 63.9  \\
			YOLOv5x & 476  & 87.3M  & 69.7 & 64.0 \\
			\bottomrule[1.2pt]
	\end{tabular}}
\vspace{-0.1in}
\end{table}
where the true positive (TP) and true negative (TN) denote correct prediction, and the false positive (FP) and false negative (FN) denote incorrect outcome. The precision and recall are correlated with the commission and omission errors, respectively. The mAP is a comprehensive indicator obtained by averaging AP values, which uses an integral method to calculate the area enclosed by the Precision-Recall curve and coordinate axis of all categories. Hence, the mAP can be calculated by
\begin{equation}
	mAP=\frac{AP}{N}=\frac{\int_{0}^{1}{p(r)dr}}{N},
\end{equation}
where  $p$ denotes Precision, $r$ denotes Recall, and $N$ is the number of categories. 

GFOLPs (Giga Floating-point Operations Per Second) and parameter size are used to measure the model complexity and computation cost. In addition, PSNR and SSIM are used for image quality evaluation of the SR branch. Generally, higher PSNR values and SSIM values represent the better quality of the generated image.

\subsection{Ablation Study}
\label{sec: Ablation Study}

First of all, we verify the effectiveness of our proposed method by designing a series of ablation experiments which are conducted on the first fold of the validation set.

\subsubsection{\textbf{Validation of the Baseline Framework}} 
In Table \ref{modelsize}, the model size and inference ability of different base frameworks are evaluated in terms of the number of layers, parameter size and GFLOPs. The detection performances of those models are measured by $\text{mA}{{\text{P}}_{\text{50}}}$, i.e., detection metric of mAP at IOU (Intersection over Union) = 0.5. Although YOLOv4 achieves the best detection performance, it has 169 more layers than YOLOv5s (393 vs. 224), its parameter size (params) is 7.4 times larger than that of YOLOv5s (52.5M vs. 7.1M), and its GFLOPs is 7.2 times higher than that of YOLOv5s (38.2 vs. 5.3). With respect to YOLOv5s, although its mAP is slightly lower than those of YOLOv4 and YOLOv5m, its number of layers, parameter size and GFLOPs are much smaller than those of other models. Therefore, it is easier to deploy YOLOv5s on board to achieve real-time performance in practical applications. The above fact verifies the rationality of YOLOv5s as the baseline detection framework.

\begin{table}[htpb]
	\small
	\centering
	\caption{Influence of Removing the Focus Module in the Network on the First Fold of the VEDAI Validation Set.}
	\label{Focus}
	\scalebox{0.95}{
	\begin{tabular}{c|c|c|c|c}
		\toprule[1.2pt]
		\multicolumn{2}{c|}{\textbf{Method}}  & \textbf{Params} $\downarrow$& \textbf{GFLOPs}$\downarrow$  & \textbf{$\text{mA}{{\text{P}}_{\text{50}}}$} $\uparrow$ \\
		\midrule
		\multirow{2}{*}{YOLOv5s} & Focus    & 7.0739M  & 5.3  & 62.2 \\
		& noFocus  & 7.0705M  & 20.4 
		& \textbf{69.5} \textcolor{blue}{(+7.3)}\\
		\midrule
		\multirow{2}{*}{YOLOv5m} & Focus   & 21.0677M  & 16.1 & 64.5\\
		& noFocus & 21.0625M    & 63.6 
		& \textbf{72.2} \textcolor{blue}{(+7.7)} \\
		\midrule
		\multirow{2}{*}{YOLOv5l} & Focus  & 46.6406M   & 36.7  & 63.7 \\
		& noFocus  & 46.6337M  & 145.0 
		& \textbf{72.5} \textcolor{blue}{(+8.8)}  \\
		\midrule
		\multirow{2}{*}{YOLOv5x} & Focus  & 87.2487M  & 69.7 & 64.0\\
		& noFocus  & 87.2400M  & 276.6 
		& \textbf{69.2} \textcolor{blue}{(+5.2)} \\
		\bottomrule[1.2pt]
	\end{tabular}}
\vspace{-0.1in}
\end{table}
\begin{table}[htpb]
	\small
	\centering
			\caption{The Comparison Result of Pixel-level and Feature-level Fusions in YOLOv5s (noFocus) for Multimodal Dataset on the First Fold of the VEDAI Validation Set.}
			\label{fusioncomparison}
			\scalebox{0.95}{
				\begin{tabular}{c|c|c|c|c}
					\toprule[1.2pt]
					\multicolumn{2}{c|}{\textbf{Method}}  & \textbf{Params} $\downarrow$ & \textbf{GFLOPs} $\downarrow$& \textbf{$\text{mA}{{\text{P}}_{\text{50}}}$} $\uparrow$ \\
					\midrule
					\multirow{2}{*}{\begin{tabular}[c]{@{}c@{}}\textbf{Pixel-level} \\   \textbf{Fusion}\end{tabular}} &Concat & \textbf{7.0705M}   & \textbf{20.37} & 69.5   \\
				    & MF & 7.0897M  & 21.67  & \textbf{70.3} \\
					\midrule
					\multirow{4}{*}{\begin{tabular}[c]{@{}c@{}}\textbf{Feature-level} \\ \textbf{Fusion}\end{tabular}} & Fusion1   & 7.0887M   & 21.76   & 66.0  \\
					& Fusion2   & 7.0744M   & 22.04   & 68.5  \\
					& Fusion3   & 7.1442M   & 24.22   & 64.8  \\
					& Fusion4   & 7.0870M   & 24.50   & 63.8  \\  
				    \midrule
				    \multicolumn{2}{c|}{\begin{tabular}[c]{@{}c@{}}\textbf{Multistage Feature-level}\\ \textbf{Fusion}\end{tabular}}    & 7.7545M   & 34.56   & 59.3  \\
				   \bottomrule[1.2pt]
				    
			\end{tabular}}
			\vspace{-0.1in}
\end{table}

\subsubsection{\textbf{Impact of Removing Focus Module}}
As presented in Section~\ref{Sec:Focus Removal}, the Focus module reduces the resolution of input images, which imposes encumbrance on the detection performance of small objects in RSI. To investigate the influence of the Focus module, we conduct experiments on the four YOLOv5 network frameworks: YOLOv5s, YOLOV5m, YOLOv5l, and YOLOv5x. Note that the results here are collected after the concatenation pixel-level fusion of RGB and IR modalities. As listed in Table \ref{Focus}, after removing the Focus module, we observe a noticeable improving in the detection performance of YOLOv5s (62.2\%$\to$69.5\% in $\text{mA}{{\text{P}}_{\text{50}}}$), YOLOv5m (64.5\%$\to$72.2\%), YOLOV5l (63.7\%$\to$72.5\%), YOLOv5x (64.0\%$\to$69.2\%). This is because by removing the Focus module, not only can the resolution degradation be avoided, but also the spatial interval information be retained for small objects in RSI, thereby reducing the missing errors of object detection. Generally, removing the Focus module brings more than 5\% improvement in the detection performance ($\text{mA}{{\text{P}}_{\text{50}}}$) of the whole frameworks. 

Meanwhile, we notice that the above removal increases the inference computation cost (GFLOPs) in YOLOv5s (5.3$\to$20.4), YOLOv5m (16.1$\to$63.6), YOLOV5l (36.7$\to$145), YOLOv5x (69.7$\to$276.6). However, the GFLOPs of YOLOv5s-noFocus (20.4) is smaller than those of YOLOv3 (52.8), YOLOv4 (38.2), and YOLOrs (46.4), as shown in Table \ref{modelsize}. The parameters of these models are slightly reduced after removing the Focus module. In summary, in order to retain the resolution to better detect smaller objects, priority shall be given to the detection accuracy, for which the convolution operation is adopted to replace the Focus module.

\begin{figure}[htpb]
	\centering
	\includegraphics[scale=0.74]{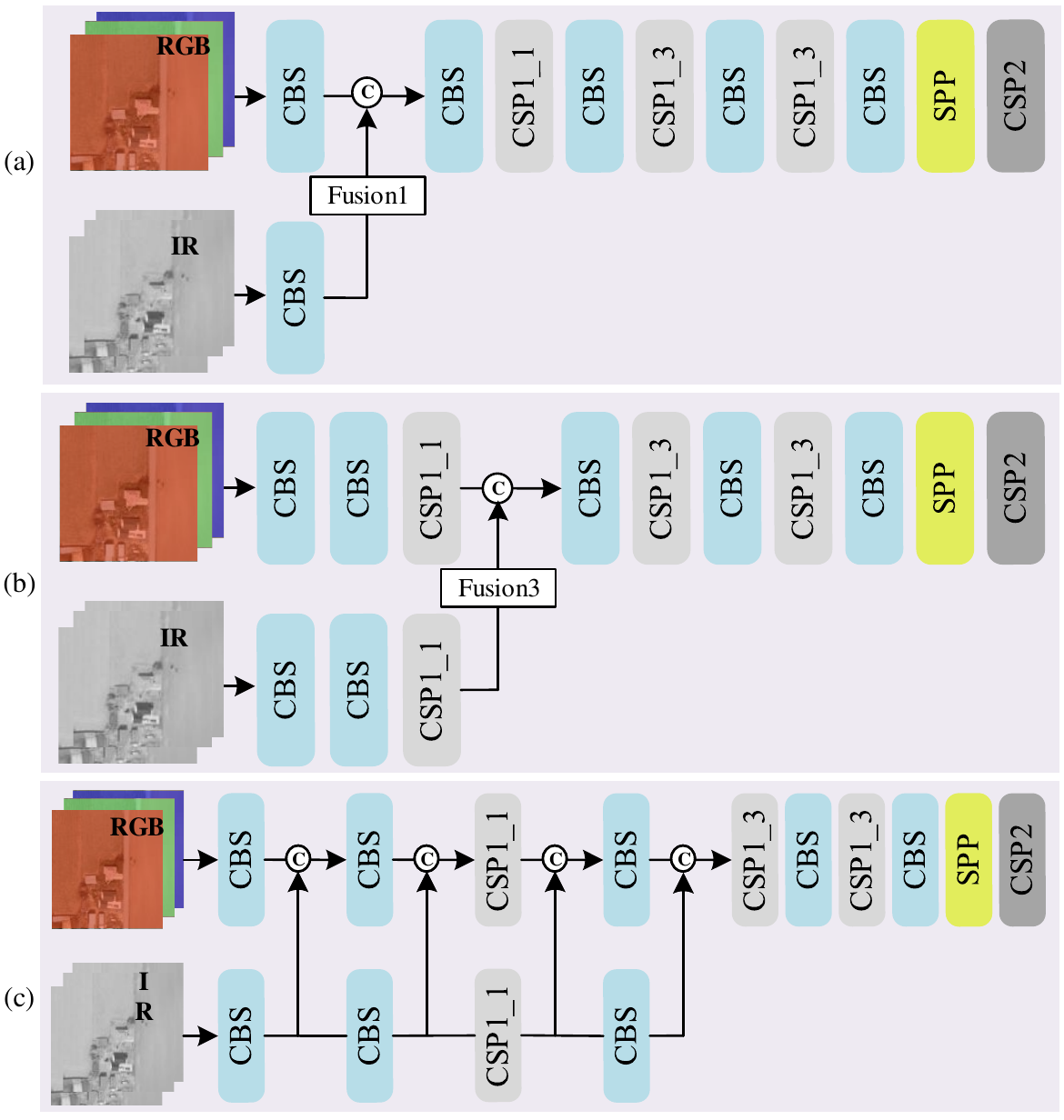}
	\centering
	\caption{The feature-level fusion of different blocks in the latent layers. Fusion-n represents the concatenation fusion operation performed in the n-th blocks. (a) and (b) is feature-level fusion and (c) is multistage feature-level fusion.}
	\vspace{-0.1in}
	\label{midfusion}
\end{figure}

\subsubsection{\textbf{Comparison of Different Fusion Methods}}
To evaluate the influence of the devised fusion methods, we compare five fusion results on YOLOv5-noFocus, as presented in Section~\ref{subsec:fusion}. As shown in Fig \ref{midfusion}, fusion1, fusion2, fusion3, and fusion4 represent the concatenation fusion operation performed in the first, second, third, and fourth blocks, respectively. The IR image is expanded to three bands in feature-level fusion to obtain the features which have equal channels for the two modes. The final result is listed in TABLE \ref{fusioncomparison}. The parameter size, GFLOPs, $\text{mA}{{\text{P}}_{\text{50}}}$ of pixel-level fusion with concatenation operation are 7.0705M, 20.37 and 69.5\%,  and these of the pixel-level fusion with MF module are 7.0897M, 21.67 and 70.3\% which are the best among all the compared methods. There are some reasons why the model parameters of the feature-level fusions are close to the pixel-level fusion. First, the feature-level fusion is completed in the latent layers rather than the whole two separate models. Second, the modules before the concatenation fusion are different, making the different fusion channels cause different parameters. However, it can be proved that calculation cost is in creased with the layer of fusion becomes deeper. In addition, we compare the multistage feature-level fusion (shown in Fig. 7 (c)) with the proposed pixel-level fusion. As shown in TABLE \ref{fusioncomparison} the accuracy of multistage feature-level fusion is only 59.3\% $\text{mA}{{\text{P}}_{\text{50}}}$ lower than that of pixel-level fusion, while its computation cost is 34.56 GFLOPs with 7.7545M parameters, which is higher than that of pixel-level fusion. These findings suggest that innovative pixel-level fusion methods are more effective than multistage shallow feature-level fusion. Because the multiple stages of fusion can lead to the accumulation of redundant information. The above results suggest that pixel-level fusion can accurately detect objects while reducing the computation. Our proposed MF fusion can improve detection accuracy with some computation costs. Overall, the proposed method only uses pixel-level fusion to contain the lower computation cost.

\begin{table}[htbp]
	\small
	\centering
	\setlength{\tabcolsep}{1.2mm}{
			\caption{The Influence of Different Resolutions for Input Image on Network Performance on the First Fold of the VEDAI Validation Set.}
			\label{resolution}
			\begin{tabular}{c|c|c|c|c|c}
				\toprule[1.2pt]
				\textbf{Method}  
				& \multicolumn{1}{c|}{\begin{tabular}[c]{@{}c@{}} \textbf{Train-Val} \\ \textbf{Size} \end{tabular}}
				& \multicolumn{1}{c|}{\begin{tabular}[c]{@{}c@{}} \textbf{Test} \\ \textbf{Size} \end{tabular}}
				& \textbf{Params} $\downarrow$ & \textbf{GFLOPs} $\downarrow$  & \textbf{$\text{mA}{{\text{P}}_{\text{50}}}$} $\uparrow$ \\
				\midrule
				\textbf{\multirow{4}{*}{YOLOv5s}}  & \multirow{2}{*}{512}   & 512  & {7.0739M} & \textbf{5.3}  & \textbf{62.2} \\
				&  & 1024 & {7.0739M} & 21.3 & 10.6 \\ 
				\cmidrule{2-6} 
				& \multirow{2}{*}{1024} & 1024 & {7.0739M} & 21.3 & \textbf{77.7} \\
				&   & 512 & {7.0739M} & \textbf{5.3}  & 48.2 \\
				\midrule
				\textbf{\multirow{4}{*}{\begin{tabular}[c]{@{}c@{}}YOLOv5s\\ (noFocus)\end{tabular}}} & \multirow{2}{*}{512}   & 512  & {7.0705M} & \textbf{20.4}  & \textbf{69.5} \\
				&  & 1024 & {7.0705M} & 81.5 & 13.4 \\ 
				\cmidrule{2-6} 
				& \multirow{2}{*}{1024} & 1024 & {7.0705M} & 81.5 & \textbf{79.3} \\
				&  & 512 & {7.0705M} & \textbf{20.4} & 62.9 \\ 
				\midrule
				\textbf{ \begin{tabular}[c]{@{}c@{}}YOLOv5s\\ (noFocus)\\ +SR\end{tabular}}  & 512  & 512  & {7.0705M} & \textbf{20.4} & \textbf{78.0} \\
				\bottomrule[1.2pt]  
		\end{tabular}}
\vspace{-0.1in}               
\end{table}
\begin{table*}[htbp]
	\small
	\centering
	\caption{The Ablation Experiment Results about the Influence of SR Branch on Detection Performance on the First Fold of the VEDAI Validation Set.}
	\label{ablation}
	\begin{tabular}{c|c|c|c|c|c|c|c|c|c}	
		\toprule[1.2pt]
		\multirow{8}{*}{\textbf{\begin{tabular}[c]{@{}c@{}}YOLOv5s\\ (noFocus)\end{tabular}} } &
		\textbf{\begin{tabular}[c]{@{}c@{}}Branch \\ \end{tabular}} & 
		\textbf{\begin{tabular}[c]{@{}c@{}}Small-scale\\ Detector\end{tabular}} &
		\textbf{\begin{tabular}[c]{@{}c@{}}Decoder\\ (EDSR)\end{tabular}} &
		\textbf{L1 Loss} &
		\textbf{Params} $\downarrow$&
		\textbf{GFLOPs} $\downarrow$&
		\textbf{$\text{mA}{{\text{P}}_{\text{50}}}$} $\uparrow$&
		\textbf{PSNR} $\uparrow$&
		\textbf{SSIM} $\uparrow$\\
		\midrule
		& Upsample & &              &              & 7.0705M & 20.37G & 76.2   & - & - \\
		& SR & &              &              & 7.0705M & 20.37G & 78.0   & - & - \\
		& SR& $\checkmark$ &              &              & 4.8259M & 16.68G & 79.0   & 23.811 & 0.602 \\
		& SR  & $\checkmark$ & $\checkmark$ &              & 4.8259M & 16.68G & 79.9 & 23.902 & 0.604 \\
		&  SR  & $\checkmark$ & $\checkmark$ & $\checkmark$ & \textbf{4.8259M} & \textbf{16.68G} & \textbf{80.9} & \textbf{26.203} & \textbf{0.659} \\
		\bottomrule[1.2pt]   
	\end{tabular}
\vspace{-0.1in}
\end{table*}

\begin{table}[htpb]
	\small
	\centering
	{
	\caption{The Effective Validation of  the Super Resolution branch for the Different  Baseline on on the First Fold of the VEDAI Validation Set.}
	\label{ablationsr}
	\begin{tabular}{c|c|c|c|c}
		\toprule[1.2pt]
        \textbf{Method} & \textbf{Layers} & \textbf{Params} $\downarrow$ & \textbf{GFLOPs} $\downarrow$ & \textbf{$\text{mA}{{\text{P}}_{\text{50}}}$} $\uparrow$ \\
		\midrule
		\textbf{YOLOv3}     & 270 & 61.5M & 52.8 & 62.6 \\
		\textbf{YOLOv3+SR}  & 270 & 61.5M & 52.8 & \textbf{71.8} \\
		\midrule
		\textbf{YOLOv4}     & 393 & 52.5M & 38.2 & 65.7 \\
		\textbf{YOLOv4+SR}  & 393 & 52.5M & 38.2 & \textbf{69.0} \\
		\midrule
		\textbf{YOLOv5s}    & 224 & 7.1M  & 5.3  & 62.2 \\
		\textbf{YOLOv5s+SR} & 224 & 7.1M  & 5.3  & \textbf{64.4}   \\ 
		\bottomrule[1.2pt]                                     
	\end{tabular}}
\vspace{-0.1in}
\end{table}

\subsubsection{\textbf{Impact of High Resolution}}
We compare different training and test modes to explore more possibilities in terms of the input image resolution in TABLE \ref{resolution}. First, we compare cases where the image resolutions of the training set and test set are the same.
By comparing the result of YOLOv5s, the detection metric $\text{mA}{{\text{P}}_{\text{50}}}$ is improved from 62.2\% to 77.7\%, causing 15.5\% increase  when the image size is doubled from 512 to 1024. Similarly, YOLOv5s-noFocus (1024) outperforms YOLOv5s-noFocus (512) by 9.8\% $\text{mA}{{\text{P}}_{\text{50}}}$ score (79.3\% vs. 69.5\%). The mean recall and mean precision increase simultaneously, suggesting that ensuring resolution reduces the commission and omission errors in object detection. Based on the above analysis, we argue that the characteristics of HR significantly influence the final performance of object detection. However, it is noteworthy that maintaining an HR input image of the network introduces a certain amount of calculation. The GFLOPs with a size of 1024 (high resolution) is higher than that with 512 (low resolution) in both YOLOv5s (21.3 vs. 5.3) and YOLOv5s-noFocus (81.5 vs. 20.4). 

As shown in Table \ref{resolution}, the use of different sizes of the image during the training process (train size) and the test process (test size) results in the score reduction of $\text{mA}{{\text{P}}_{\text{50}}}$, i.e.,  (10.6\% vs. 62.2\%), (48.2\% vs. 77.7\%), (13.4\% vs. 69.5\%) and (62.9\% vs. 79.3\%). This may attribute to the inconsistent scale of objects in the test process and in the training process, where the size of the predicted bounding box is not suitable for the objects of test images anymore.

Finally, the $\text{mA}{{\text{P}}_{\text{50}}}$ of YOLOv5s-noFocus+SR is close to the YOLOv5-noFocus  HR (1024) one (78.0\% vs. 79.3\%), and the GFOLPs is equal to that of YOLOv5-noFocus LR (512)  one (20.4 vs. 20.4). Our proposed network decreased the resolution of input images in the test process to reduce computation and maintain accuracy by remaining the identical resolution of the training and testing data, thereby highlighting the advantage of the proposed SR branch.

\subsubsection{\textbf{Impact of Super Resolution Branch}}

Some ablation experiences about the SR branch are completed in Table \ref{ablation}. Compared with the upsampling operation, the YOLOv5s (noFocus) added super resolution network shows favorable performance and gets $\text{mA}{{\text{P}}_{\text{50}}}$ 1.8\% better than upsampling operation. The SR network is a learnable upsampling method with a more vital reconstruction ability that can help the feature extraction in the backbone for detection.  We deleted the PANet structure and two detectors, which are responsible for enhancing middle-scale and large-scale target detection because the objects RSI datasets, such as VEDAI are on the small scale and can be detected with the small-scale detector. When we only use one detector, the number of parameters (7.0705M vs. 4.8259M) and GFLOPs (20.37 vs. 16.68) can be decreased, and the detection accuracy can be increased (78.0\%  vs 79.0\% ). When we utilize the EDSR network (rather than three ordinal deconvolutions) as Decoder and L1 loss (rather than L2 loss) as SR loss function in the SR branch, which is powerful in the SR task, not only the performance of SR is improved but also the performance of the detection network enhanced meantime because the SR branch helps the detection network to extract more effective and superior features in the backbone, accelerating the convergence of the detection network and thus improving the performance of the detection network. The performance of super resolution and object detection is complementary and cooperative. 

Table \ref{ablationsr} shows the favorable accuracy-complexity tradeoff of the SR branch. At the different baselines, the influence of the SR branch on object detection is positive. Compared with bare baseline, baseline added super resolution shows favorable performance: YOLOv3+SR performs $\text{mA}{{\text{P}}_{\text{50}}}$ 9.2\% better than YOLOv3, YOLOv4+SR is $\text{mA}{{\text{P}}_{\text{50}}}$ 3.3\% better than YOLOv4, YOLOv5s+SR performs $\text{mA}{{\text{P}}_{\text{50}}}$ 2.2\% better than YOLOv5s. Notably, super resolution can be removed in the inference stage. Hence no extra parameters and computation costs are introduced, which is impressive considering that the SR branch does not require a lot of manpower to refine the design of the detection network. The SR branch is general and extensible and can be utilized in the existing fully convolutional network (FCN) framework.

\begin{figure*}[htpb]
	\centering
	\includegraphics[scale=0.85]{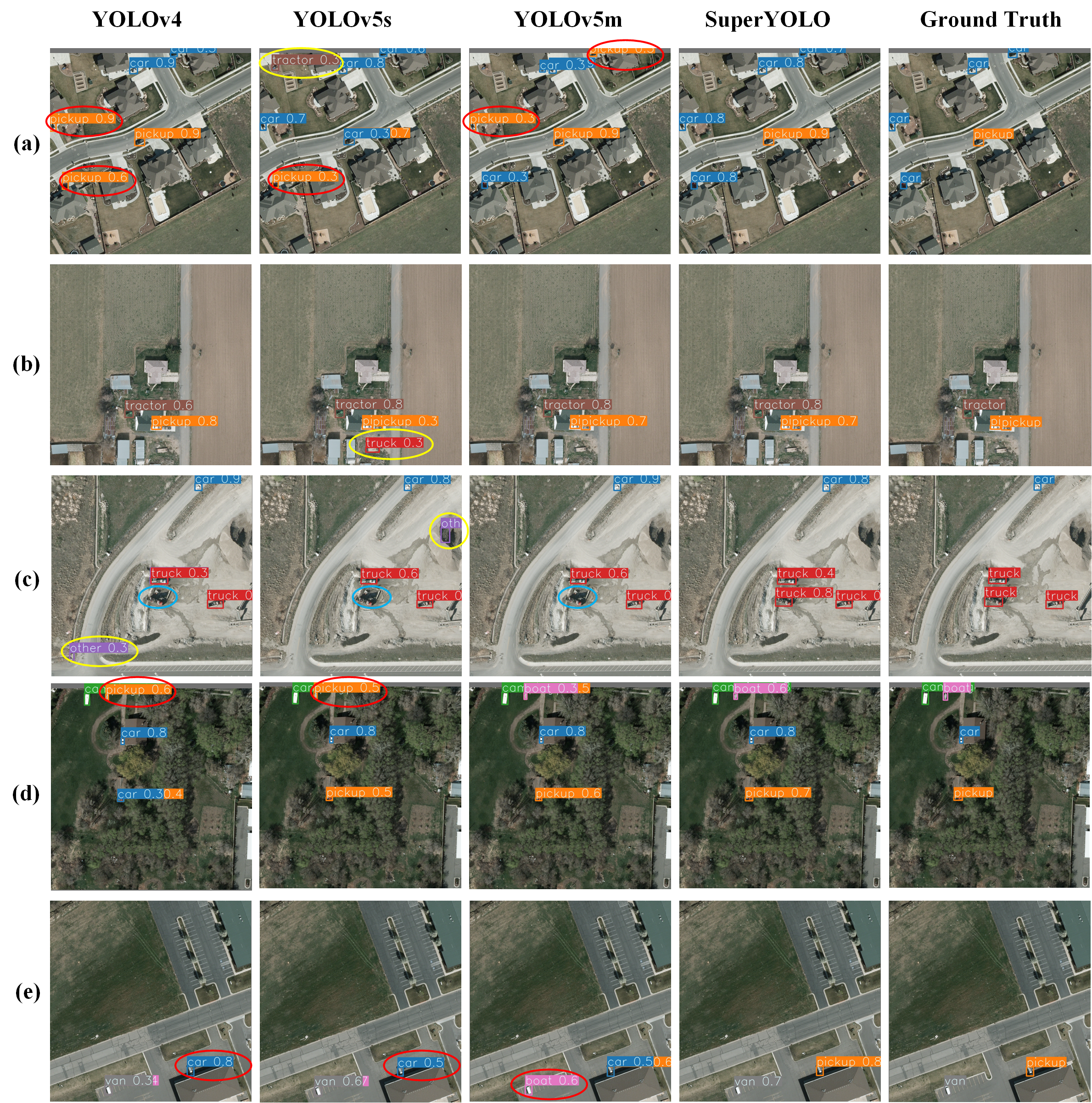}
	\centering
	\caption{Visual results of object detection using different methods involving YOLOv4, YOLOv5s, YOLOv5m and the proposed SuperYOLO. The red cycles represent the False Alarms, the yellow ones denote the False Positive detection results and the blue ones are False Negative detection results. (a)-(e) is the different images in the VEDAI dataset.}
	\vspace{-0.1in}
	\label{results}
\end{figure*}
\begin{table*}[htpb]
	\small
	\renewcommand{\arraystretch}{1.2}
	\centering
	\setlength{\tabcolsep}{1mm}{
	\caption{Class-wise Average Precision AP, Mean Average Precision {\upshape$\text{mA}{{\text{P}}_{\text{50}}}$}, Parameters and GFLPs  for Proposed SuperYOLO, YOLOv3, YOLOv4, YOLOv5s-x, YOLOrs, YOLO-Fine and YOLOFusion Including Unimodal and Multimodal Configurations on VEDAI Dataset. * Represents Using Pre-trained Weight.}
	\label{comparedresult}
	\begin{tabular}{c|c|c|c|c|c|c|c|c|c|c|c|c}
		\toprule[1.2pt]           		\multicolumn{2}{c|}{\textbf{Method}}           & \textbf{Car}            & \textbf{Pickup}         & \textbf{Camping}        & \textbf{Truck}         & \textbf{Other}          & \textbf{Tractor}        & \textbf{Boat}           & \textbf{Van}            & \textbf{$\text{mA}{{\text{P}}_{\text{50}}}$}  $\uparrow$ & \textbf{Params} $\downarrow$ & \textbf{GFLOPs} $\downarrow$ \\
		\midrule
		\textbf{\multirow{3}{*}{YOLOv3 \cite{redmon2018yolov3}}}& IR         & 80.21          & 67.03          & 65.55          & 47.78          & 25.86          & 40.11          & 32.67          & 53.33          & 51.54    &\textbf{61.5351M} &\textbf{49.55}      \\
		& RGB        & 83.06          & 71.54          & \textbf{69.14} & 59.30          & \textbf{48.93} & \textbf{67.34} & 33.48          & 55.67          & 61.06       &\textbf{61.5351M} &\textbf{49.55}   \\
		& Multi & \textbf{84.57} & \textbf{72.68} & 67.13          & \textbf{61.96} & 43.04          & 65.24          & \textbf{37.10} & \textbf{58.29} & \textbf{61.26} & 61.5354M & 49.68\\
		\midrule
		\textbf{\multirow{3}{*}{YOLOv4 \cite{bochkovskiy2020yolov4}}}  & IR         & 80.45          & 67.88          & 68.84          & 53.66          & 30.02          & 44.23          & 25.40          & 51.41          & 52.75       &\textbf{52.5082M} &\textbf{38.16}   \\
		& RGB        & 83.73          & \textbf{73.43} & 71.17          & 59.09          & \textbf{51.66} & 65.86          & \textbf{34.28} & \textbf{60.32} & 62.43         &\textbf{52.5082M} &\textbf{38.16} \\
		& Multi & \textbf{85.46} & 72.84          & \textbf{72.38} & \textbf{62.82} & 48.94          & \textbf{68.99} & \textbf{34.28} & 54.66          & \textbf{62.55} & 52.5085M  & 38.23 \\
		\midrule
		\textbf{\multirow{3}{*}{YOLOv5s \cite{yolov5}}} & IR         & 77.31          & 65.27          & 66.47          & 51.56          & 25.87          & 42.36          & 21.88          & 48.88          & 49.94       & \textbf{7.0728M}  & \textbf{\textcolor{blue}{5.24}}   \\
		& RGB        & 80.07          & 68.01          & 66.12          & 51.52          & 45.76          & \textbf{64.38} & 21.62          & 40.93          & 54.82          & \textbf{7.0728M}  & \textbf{\textcolor{blue}{5.24}} \\
		& Multi & \textbf{80.81} & \textbf{68.48} & \textbf{69.06} & \textbf{54.71} & \textbf{46.76} & 64.29          & \textbf{24.25} & \textbf{45.96} & \textbf{56.79} & 7.0739M        & 5.32 \\
		\midrule
		\textbf{\multirow{3}{*}{YOLOv5m \cite{yolov5}} }& IR         & 79.23          & 67.32          & 65.43          & 51.75          & 26.66          & 44.28          & 26.64          & 56.14          & 52.19    &\textbf{21.0659M} &\textbf{16.13}      \\
		& RGB        & 81.14          & 70.26          & 65.53          & 53.98          & \textbf{46.78} & \textbf{66.69} & \textbf{36.24} & 49.87          & 58.80     &\textbf{21.0659M} &\textbf{16.13}     \\
		& Multi & \textbf{82.53} & \textbf{72.32} & \textbf{68.41} & \textbf{59.25} & 46.20          & 66.23          & 33.51          & \textbf{57.11} & \textbf{60.69}  & 21.0677M  & 16.24 \\
		\midrule
		\textbf{\multirow{3}{*}{YOLOv5l \cite{yolov5}}} & IR         & 80.14          & 68.57          & 65.37          & 53.45          & 30.33          & 45.59          & 27.24          & \textbf{61.87} & 54.06     &\textbf{46.6383M} &\textbf{36.55}     \\
		& RGB        & 81.36          & 71.70          & 68.25          & 57.45          & 45.77          & \textbf{70.68} & 35.89          & 55.42          & 60.81    &\textbf{46.6383M} &\textbf{36.55}      \\
		& Multi & \textbf{82.83} & \textbf{72.32} & \textbf{69.92} & \textbf{63.94} & \textbf{48.48} & 63.07          & \textbf{40.12} & 56.46          & \textbf{62.16} &46.6406M &36.70 \\
		\midrule
		\textbf{\multirow{3}{*}{YOLOv5x \cite{yolov5}} }& IR         & 79.01          & 66.72          & 65.93          & 58.49          & 31.39          & 41.38          & 31.58          & 58.98          & 54.18     &\textbf{87.2458M} &\textbf{69.52}     \\
		& RGB        & 81.66          & 72.23          & 68.29          & 59.07          & 48.47          & 66.01          & \textbf{39.15} & \textbf{61.85} & 62.09        &\textbf{87.2458M} &\textbf{69.52}   \\
		& Multi & \textbf{84.33} & \textbf{72.95} & \textbf{70.09} & \textbf{61.15} & \textbf{49.94} & \textbf{67.35} & 38.71          & 56.65          & \textbf{62.65} & 87.2487M &69.71\\
		\midrule
		\textbf{\multirow{3}{*}{YOLOrs \cite{9273212}} }& IR         & 82.03          & 73.92         & 63.80          & \textbf{54.21 }        & 43.99          & 54.39          & \textbf{21.97 }         & 43.38          & 54.71     &- &-     \\
		& RGB        & \textbf{85.25}          & 72.93          & \textbf{70.31}          & 50.65         & 42.67          & \textbf{76.77}          & 18.65 & 38.92 & 57.00        &- &-  \\
		& Multi & 84.15 & \textbf{78.27} & 68.81 & 52.60 & \textbf{46.75} & 67.88 & 21.47         & \textbf{57.91}         & \textbf{59.73} & - &-\\
		\midrule
		\textbf{\multirow{2}{*}{YOLO-Fine \cite{pham2020yolo}} }& IR       & 76.77          & 74.35         & 64.74         & 63.45          & \textbf{45.04}          & \textbf{78.12}          & \textbf{70.04} & \textbf{77.91} & 68.18        &- &-   \\
		& RGB         & \textbf{79.68}         & \textbf{74.49}        & \textbf{77.09}          & \textbf{80.97}          & 37.33          & 70.65          & 60.84          & 63.56         & \textbf{68.83}     &- &-     \\
		\midrule
		\textbf{\multirow{3}{*}{YOLOFusion* \cite{qingyun2022cross}} } & IR        & 86.7          & 75.9          &66.6         & 77.1         & 43.0          & 62.3          & 70.7 & \textbf{84.3} & 70.8       &- &-  \\
		& RGB       & 91.1          & 82.3         & 75.1          & \textbf{78.3}        & 33.3          & \textbf{81.2}          & \textbf{71.8}         & 62.2          &  71.9    &- &-     \\
		& Multi & \textbf{91.7} & \textbf{85.9} & \textbf{78.9} & 78.1 & \textbf{54.7}& 71.9 & 71.7         &  75.2    & \textbf{\textcolor{blue}{75.9}}  & 12.5M &-\\
		\midrule
		\textbf{\multirow{3}{*}{SuperYOLO}} & IR         & 87.90          & 81.39          & 76.90          & 61.56          & 39.39          & 60.56          & 46.08          & 71.00 & 65.60          & \textbf{\textcolor{blue}{4.8256M}}  & \textbf{16.61}\\
		& RGB        & 90.30          & 82.66 & 76.69          & 68.55          & 53.86          & 79.48 & 58.08          & 70.30          & 72.49         & \textbf{\textcolor{blue}{4.8256M}}  & \textbf{16.61}\\
		& Multi & \textbf{91.13} & \textbf{85.66}        & \textbf{79.30} & \textbf{70.18} & \textbf{57.33} & \textbf{80.41}        & \textbf{60.24} & \textbf{76.50}          & \textbf{\textcolor{blue}{75.09}} & 4.8451M  & 17.98 \\
		\bottomrule[1.2pt]
	\end{tabular}}
\vspace{-0.1in}
\end{table*}

\subsection{Comparisons with Previous Methods}
The visual detection results of the compared YOLO methods and SuperYOLO are shown in Fig. \ref{results}, for a diverse set of scenes. It can be observed that SuperYOLO can accurately detect those objects that are not detected, or predicted into a wrong category or with uncertainty, in YOLOv4, YOLOv5s, and YOLOv5m. The objects in RSIs are challenging to detect on small scales. In particular, \textbf{Pickup} and \textbf{Car} or \textbf{Van} and \textbf{Boat} are easily confused in the detection process due to their similarities. Hence, improving the detection classification is of essential necessity in object detection tasks except for location detection, which can be accomplished by the proposed SuperYOLO with better performance. 

TABLE \ref{comparedresult} summarizes the performance of the \textbf{YOLOv3 \cite{redmon2018yolov3}}, \textbf{YOLOv4  \cite{bochkovskiy2020yolov4}}, and \textbf{YOLOv5s-x} \cite{yolov5} \textbf{YOLOrs } \cite{9273212}, \textbf{YOLO-Fine} \cite{pham2020yolo}, \textbf{YOLOFusion} \cite{qingyun2022cross} and our proposed \textbf{SuperYOLO}. Note that the AP scores of multimodal modes are significantly higher than those of unimodal (RGB or IR) modes for most classes. The overall $\text{mA}{{\text{P}}_{\text{50}}}$ of multimodal (multi) modes outperforms those of RGB or IR modes. These results confirm that multimodal fusion is an effective and efficient strategy for object detection based on information complementation between multimodal inputs. However, it should be noted that the slight increase in parameters and GFLOPs with multimodal fusion reflects the necessity of choosing pixel-level fusion rather than feature-level fusion.

It is obvious that the SuperYOLO achieves higher $\text{mA}{{\text{P}}_{\text{50}}}$ than the other frameworks except for YOLOFusion. The results of YOLOFusion are slightly better than SuperYOLO, as YOLOFusion uses pre-trained weight which is trained on MS COCO \cite{lin2014microsoft}. However, its parameter count is approximately three times that of SuperYOLO. The performance of YOLO-Fine is good on a single modality, but it lacks the development of multi-modality fusion techniques. In particular, the SuperYOLO outperforms the YOLOv5x by a 12.44\% $\text{mA}{{\text{P}}_{\text{50}}}$ score in multimodal mode. Meanwhile, parameter size and GFLOPs of SuperYOLO are about 18x and 3.8x less than YOLOv5x. 

In addition, it can be noticed that the superior performance is achieved for the classes of \textbf{Car}, \textbf{Pickup}, \textbf{Tractor} and \textbf{Camping}, which have the most training instances. YOLOv5s performs superior on GFLOPs, which depends on the Focus module to slim the input image, but results in lousy detection performance, especially for small objects. The SuperYOLO performs 18.30\% $\text{mA}{{\text{P}}_{\text{50}}}$ better than YOLOv5s. Our proposed SuperYOLO shows a favorable speed-accuracy trade-off compared to the state-of-the-art models.



\subsection{Generalization to single modal remote sensing images}
At present, although there are massive multimodal images in remote sensing, the labeled dataset in object detection tasks is lacking due to the expensive cost of manually annotating. To validate the generalization of our proposed network, we compare the SuperYOLO with different one-stage or two-stage methods using data from the single modality including a large-scale Dataset for Object Detection in Aerial images (DOTA), object DetectIon in Optical Remote sensing images (DIOR), and Northwestern Polytechnical University Very-High-Resolution 10-class (NWPU VHR-10) datasets.

\textit{1) DOTA:} The DOTA dataset was proposed in 2018 for object detection of remote sensing. It contains 2806 large images and 188 282 instances, which are divided into 15 categories. The size of each original image is $4000 \times 4000$, and the images are cropped into $1024 \times 1024$ pixels with an overlap of 200 pixels in the experiment. We select half of the original images as the training set, 1/6 as the validation set, and 1/3 as the testing set.  The size of the image is fixed to $512 \times 512$.

\textit{2) NWPU VHR-10:} The dataset of NWPU VHR-10 was proposed in 2016. It contains 800 images, of which 650 pictures contain objects, so we use 520 images as the training set and 130 images as the testing set. The dataset contains 10 categories, and the size of the image is fixed to $512 \times 512$. 

\textit{3) DIOR:} The DIOR dataset was proposed in 2020 for the task of object detection, which involves 23 463 images and 192 472 instances. The size of each image is $800 \times 800$. We choose 11 725 images as the training set and 11 738 images as the testing set. The size of the image is fixed to $512 \times 512$

\begin{table*}[htpb]
	\small
	\renewcommand{\arraystretch}{1.2}
	\centering
	\setlength{\tabcolsep}{1.6mm}{
		\caption{Performance of Different Algorithms on DOTA, NWPU and DOTA Testing Set.}
		\label{otherdataset}
		\begin{tabular}{c|ccc|ccc|ccc}
			\toprule[1.2pt]
            & \multicolumn{3}{c|}{\textbf{DOTA-v1.0}}         & \multicolumn{3}{c|}{\textbf{NWPU}}                    & \multicolumn{3}{c}{\textbf{DIOR}}         \\
            \midrule
			\textbf{Method}         & \textbf{$\text{mA}{{\text{P}}_{\text{50}}}$}          & \textbf{Params(M)}  & \textbf{GFLOPs }        & \textbf{$\text{mA}{{\text{P}}_{\text{50}}}$ }               & \textbf{Params(M)}  & \textbf{GFLOPs}         & \textbf{$\text{mA}{{\text{P}}_{\text{50}}}$ }    & \textbf{Params(M)}  & \textbf{GFLOPs}         \\
			\midrule[1.2pt]
			\textbf{Faster R-CNN \cite{ren2016faster}} & 60.64 & 60.19     & 289.25   & 77.80 & 41.17     & 127.70   & 54.10 & 60.21     & 182.20   \\
			\textbf{RetainNet \cite{lin2017focal}}    & 50.39 & 55.39     & 293.36   & 89.40 & 36.29     & 123.27   & 65.70 & 55.49     & 180.62   \\
			\textbf{YOLOv3 \cite{redmon2018yolov3}}       & 60.00 & 61.63     & 198.92   & 88.30 & 61.57     & 121.27   & 57.10 & 61.95     & 122.22   \\
			\textbf{GFL \cite{li2020generalized}}          & 66.53 & 19.13     & 159.18   & 88.80 & 19.13     & 91.73    & 68.00 & 19.13     & 97.43    \\
			\textbf{FCOS \cite{tian2019fcos}}         & 67.72 & 31.57     & 202.15   & 89.65 & 31.86     & 116.63   & 67.60 & 31.88     & 123.51   \\
			\textbf{ATSS \cite{zhang2020bridging}}         & 66.84 & 18.97     & 156.01   & 90.50 & 18.96     & 89.90    & 67.70 & 18.98     & 95.50    \\
			\textbf{MobileNetV2 \cite{sandler2018mobilenetv2}}  & 56.91 & 10.30     & 124.24   & 76.90 & 10.29     & 71.49    & 58.20 & 10.32     & 76.10    \\
			\textbf{ShuffleNet \cite{zhang2018shufflenet}}   & 57.73 & 12.11     & 142.60   & 83.00 & 12.10     & 82.17    & 61.30 & 12.12     & 87.31    \\
			\textbf{O2-DNet \cite{WEI2020268}}     & 71.10          & 209.0          & -         & -                & -          & -          & 68.3     & 209.0          & -        \\
			\textbf{FMSSD \cite{2020FMSSD}}     & \textbf{\textcolor{blue}{72.43}}         &     136.04     & -         & -                & -          & -          & 69.5     & 136.03          & -        \\
			\textbf{ARSD \cite{yang2022adaptive}}         & 68.28 & 13.08     & 68.03    & 90.92 & 11.57     & 26.65    & 70.10 & 13.10     & 41.60           \\
			\textbf{SuperYOLO} & 69.99 & \textbf{\textcolor{blue}{7.70}} & \textbf{\textcolor{blue}{20.89} }& \textbf{\textcolor{blue}{93.30}}   & \textbf{\textcolor{blue}{7.68}} & \textbf{\textcolor{blue}{20.86} }&  \textbf{\textcolor{blue}{71.82}} & \textbf{\textcolor{blue}{7.70}} & \textbf{\textcolor{blue}{20.93}} \\
			\bottomrule[1.2pt]
	\end{tabular}}
	\vspace{-0.1in}
\end{table*}

The training strategy is modified to accommodate the new dataset. The entire training process involves 150 epochs for NWPU and DIOR datasets and 100 epochs for DOTA. The batch size of the DOTA and DIOR is 16 and NWPU is 8. To verify the superiority of the SuperYOLO proposed in this paper, we selected 11 generic methods for comparison: one-stage algorithms (\textbf{YOLOv3} \cite{redmon2018yolov3},  \textbf{FCOS} \cite{tian2019fcos},  \textbf{ATSS} \cite{zhang2020bridging}, \textbf{RetainNet} \cite{lin2017focal}, \textbf{GFL} \cite{li2020generalized}); two-stage method (\textbf{Faster R-CNN} \cite{ren2016faster}); lightweight models (\textbf{MobileNetV2} \cite{sandler2018mobilenetv2} and \textbf{ShuffleNet} \cite{zhang2018shufflenet}); distillation-based methods (\textbf{ARSD} \cite{yang2022adaptive}); remote sensing designed approaches (\textbf{FMSSD} \cite{2020FMSSD} and \textbf{O2DNet} \cite{WEI2020268}).

As presented in TABLE \ref{otherdataset}, our SuperYOLO achieves the optimal detection result (69.99\%, 93.30\%, 71.82\% $\text{mA}{{\text{P}}_{\text{50}}}$) and the model parameters (7.70 M, 7.68 M, and 7.70 M) and GFLOPs (20.89, 20.86, and 20.93) are much smaller than other SOTA detectors regardless of the two-stage, one-stage, lightweight or distillation-based method. The PANet structure and three detectors are responsible for enhancing small-scale, middle-scale and large-scale target detection in consideration of the big objects such as playgrounds in these three datasets. Hence the model parameters of SuperYOLO are more than those in TABLE \ref{comparedresult}. We also compare two detectors designed for remote sensing imagery such as FMSSD \cite{2020FMSSD} and O2DNet \cite{WEI2020268}. Although these models have a close performance with our lightweight model, the huger parameters and GFLOPs seem to be a massive cost in computation resources. Hence, our model has a better balance in consideration of detection efficiency and efficacy.

\section{Conclusion and Future Work}
\label{sec:Conclusion}
In this paper, we have presented SuperYOLO, a real-time lightweight network that is built on top of the widely-used YOLOv5s to improve the detection performance of small objects in RSI.
First, we have modified the baseline network by removing the Focus module to avoid resolution degradation, through which the baseline is significantly improved and overcomes the missing error of small objects. Second, we have conducted research fusion of multimodality to improve the detection performance based on mutual information. Lastly and most importantly, we have introduced a simple and flexible SR branch facilitating the backbone to construct an HR representation feature, by which small objects can be easily recognized from vast backgrounds with merely LR input required. We remove the SR branch in the inference stage, accomplishing the detection without changing the original structure of the network to achieve the same GFOLPs. With joint contributions of these ideas, the proposed SuperYOLO achieves 75.09\% $\text{mA}{{\text{P}}_{\text{50}}}$ with lower computation cost on VEDAI dataset, which is 18.30\% higher than that of YOLOv5s, and more than 12.44\% higher than that of YOLOv5x. 

The performance and inference ability of our proposal highlight the value of SR in remote sensing tasks, paving way for the future study of multimodal object detection. Our future interests will be focusing on the design of a low-parameter mode to extract HR features, thereby further satisfying real-time and high-accuracy motivations.





\ifCLASSOPTIONcaptionsoff
\newpage
\fi
{
	\bibliographystyle{IEEEtran}
	\bibliography{reference}
}

\end{document}